\pdfoutput=1

\documentclass[11pt]{article}

\usepackage[preprint]{acl}

\usepackage{times}
\usepackage{latexsym}

\usepackage[T1]{fontenc}

\usepackage[utf8]{inputenc}

\usepackage{microtype}

\usepackage{inconsolata}

\usepackage{graphicx}

\usepackage[shortcuts = false, acronym, hyperfirst = false]{glossaries}
\usepackage{todonotes}
\usepackage{amsfonts}
\usepackage{subcaption}
\usepackage{booktabs}
\usepackage{multirow}
\usepackage{cleveref}
\usepackage{svg}
\usepackage[most]{tcolorbox}
\usepackage{url}

\usepackage{amsmath, amsfonts, amssymb}
\usepackage{booktabs} 
\usepackage{multirow} 
\usepackage{makecell}

\title{Beyond Bag-of-Patches:Internalizing Global Layout via Textual Structural Supervision for Visual Document Retrieval}
\title{Beyond Bag-of-Patches: Learning Global Layout via Textual Supervision for Late-Interaction Visual Document Retrieval}

\author{Pascal Tilli\textsuperscript{*} \\
  University of Stuttgart, \\
  Stuttgart, Germany \\
  \texttt{pascal.tilli@ims.uni-stuttgart.de} \\
  \And
  Mohsen Mesgar \\
  Bosch Center for Artificial Intelligence, \\
  Renningen, Germany \\
  \texttt{mohsen.mesgar@bosch.com} \\
  }

\newtcolorbox{promptbox}{  
  colback=gray!5,        
  colframe=gray!40,      
  boxrule=0.4pt,         
  arc=3pt,               
  left=8pt,              
  right=8pt,             
  top=6pt,               
  bottom=6pt,            
  fontupper=\ttfamily,   
  sharp corners=all,     
}

\begin{document}
\maketitle

\begingroup
\renewcommand{\thefootnote}{\textsuperscript{*}}
\footnotetext{Work done during an internship at the Bosch Center for Artificial Intelligence.}
\endgroup

\newcommand{\colpali}{\texttt{ColPali}}
\newcommand{\colbert}{\texttt{ColBERT}}
\newcommand{\colqwen}{\texttt{ColQwen2.5}}
\newcommand{\nemoretriever}{\texttt{llama-nemoretriever-colembed}}
\newcommand{\colnomic}{\texttt{colnomic-embed-multimodal}}
\newcommand{\jina}{\texttt{jina-embeddings-v4}}
\newcommand{\gbldescft}{\texttt{Gbl-Desc-FT}}
\newcommand{\cczeroshot}{\texttt{CrossContext-ZS}}
\newcommand{\ccft}{\texttt{CrossContext-FT}}
\newcommand{\nemoone}{\texttt{Nemo-1b}}
\newcommand{\nemothree}{\texttt{Nemo-3b}}
\newcommand{\colnomicthree}{\texttt{Colnomic-3b}}
\newcommand{\colnomicseven}{\texttt{Colnomic-7b}}
\newcommand{\jinaembeds}{\texttt{Jina-v4}}
\newcommand{\granitevision}{\texttt{Granite}}
\newcommand{\gptfive}{\texttt{GPT-5}}

\begin{abstract}

Visual Document Retrieval (VDR) models mostly rely on late interaction architectures, in which  documents are represented by a set of local patch embeddings and then matched against query tokens.  
While efficient, this architecture  prioritizes local similarity over global layout structure of documents to estimate relevancy between documents and query. 
 In practice, this leads to errors as relevance originates from layout structure of documents with heterogeneous layouts combining figures, tables, and text.  
We make document layout learnable without changing inference. 
We propose a multimodal encoder that augments local patch representations with a global layout embedding, trained via  textual descriptions encoding document layout information.  
Across four ViDoRe-v2 datasets, our model improves over the strongest architecturally comparable ColPali/ColQwen baseline by +2.4 nDCG@5 and +2.3 MAP@5, with statistically significant per-dataset gains over ColQwen.
\end{abstract}

\newacronym{vidorev2}{ViDoRe-v2}{Visual Document Retrieval Benchmark V2}
\newacronym{vllm}{\textsc{Vllm}}{Vision Large Language Model}
\newacronym{llm}{\textsc{llm}}{Large Language Model}
\newacronym{ocr}{OCR}{Optical Character Recognition}
\newacronym{vdr}{VDR}{Visual Document Retrieval}
\newacronym{rag}{RAG}{Retrieval‑Augmented Generation}
\newacronym{ir}{IR}{Information Retrieval}
\newacronym{tfidf}{TF-IDF}{term frequency–inverse document frequency}
\newacronym{bm25}{BM25}{Best Matching 25}
\newacronym{clip}{\textsc{Clip}}{Contrastive Language-Image Pre-training}
\newacronym{sbert}{\textsc{Sbert}}{Sentence-BERT}
\newacronym{qa}{QA}{question answering}
\newacronym{gme}{GME}{General Multimodal Embedder}
\newacronym{dse}{DSE}{Document Screenshot Embedding}
\newacronym{lora}{LoRA}{Low-Rank Adapters}
\newacronym{ndcg}{\texttt{nDCG}}{Normalized Discounted Cumulative Gain}
\newacronym{map}{\texttt{MAP}}{Mean Average Precision}
\newacronym{mteb}{\texttt{MTEB}}{Massive Text Embedding Benchmark}

\section{Introduction}
\label{sec:intro}
\begin{figure}[!t]
    \small
    \centering
    \includegraphics[width=1.0\linewidth,trim={0.7cm 3.5cm 0.7cm 3.5cm},clip]{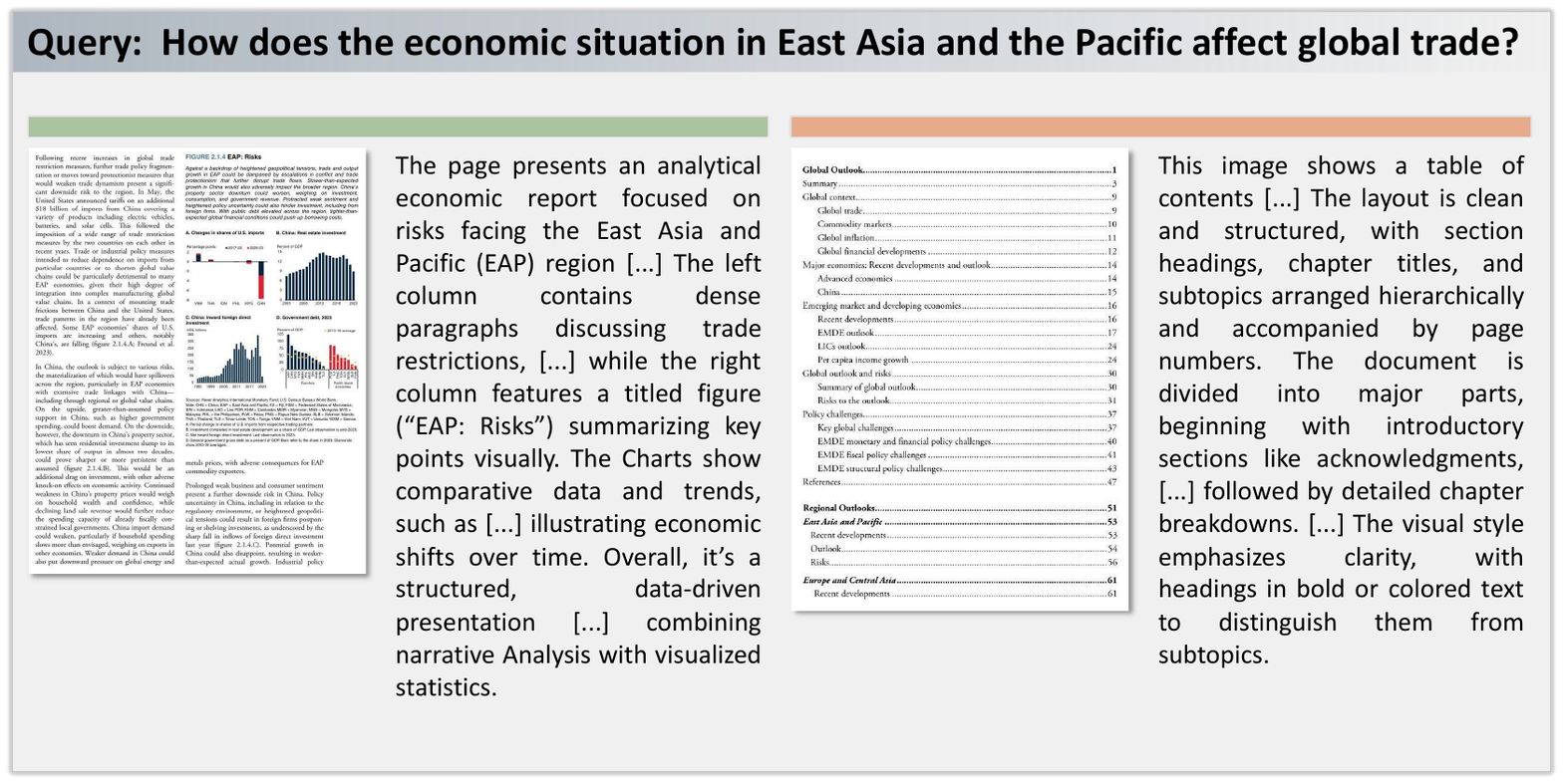}
    \caption{
        Example of a layout-intensive query. 
        The relevant page (left) combines narrative evidence with a chart region, requiring cross-region reasoning over text, figures, and layout. 
        The distractor page (right) contains superficial lexical overlap but is globally irrelevant because it is a table of contents. 
        Textual structural descriptors summarize page-level organization during training, helping the model learn global layout signals that patch-only MaxSim scoring tends to miss.
    }
    \label{fig:teaser}
\end{figure}
Document retrieval is a core component of retrieval-augmented generation (RAG), where retrieval quality directly determines factual grounding \cite{rag, rag_survey, wu-etal-2025-doc}.
Documents in practical domains exhibit heterogeneous layouts combining text, tables, and figures \cite{mmdocir, unidoc}.
This incentivize models to integrate semantically and visually diverse features. 
However, recent embedding models fail to capture document-level semantics in visually structured pages since they are primarily optimized for text-dominant corpora \cite{rag_survey, abootorabi-etal-2025-ask}.

A promising solution is \emph{visual document retrieval (VDR)} models, which treat document pages as images to avoid expensive OCR \cite{colpali, Nemoretriever, nomicembedmultimodal2025, jina-colembeds}. 
These VDR models often adopt \emph{late interaction} (a.k.a \emph{MaxSim}), representing each page as a set of patch embeddings and computing relevance via a sum of maximum similarities between query tokens and page patch tokens \cite{colbert}.  
While being effective for capturing fine-grained local evidences such as table cells, figure regions, or typographic cues,
their \emph{MaxSim} formulation ignores spatial relationships between matched regions and biases retrieval toward isolated local matches. 

In practical domains, relevance in visually rich documents depends also on \emph{page-level structure}, i.e., how textual and visual elements interact across the page. 
As illustrated in \Cref{fig:teaser}, answering the query requires linking narrative text about economic risks with corresponding figures, rather than matching either in isolation. 
This exposes a fundamental challenge as pages may contain locally relevant elements while remaining globally irrelevant. 
As a result, global layout signals, which may be partially encoded by vision encoders, remain weakly expressed in the retrieval space after late-interaction scoring. 
We argue that this is a structural limitation of patch-based retrieval and empirically show that global relevance cannot be recovered from local similarity alone.


We propose a late-interaction encoder that introduces a structural bottleneck via a \texttt{[CLS]}-style global token. 
This token is explicitly optimized to capture global information about the page and enables local patches be contextualized by page-level layout.
Rather than relying on pooling or inference-time augmentation, we train global structure directly within the retrieval space.
The core idea lies in \emph{descriptor-guided structural supervision}.
We train the global token using textual descriptors of document layout, instead of standard image--caption objectives. 
These automatically-generated descriptors summarize page-level organization and are used only during training, encouraging the model to encode structural relationships alongside local visual evidence.
It enables retrieval in layout-intensive domains where relevance depends on cross-region interactions. 
At inference, descriptors are discarded since the model indexes only the page images plus one learned global vector, preserving the standard visual retrieval pipeline with negligible overhead.


%
We experiment on the four datasets of \gls*{vidorev2}.
We compare our model with \colpali/\texttt{ColQwen} and larger models such as \nemothree{}.
To study the effect of descriptor-guided supervision, we introduce two \texttt{Cross-Context} methods as baselines. 
These methods inject textual descriptors at inference and act as an oracle for explicit access to global structure information. 
Our model improves over \colpali/\texttt{ColQwen} by \textbf{+2.4 nDCG@5} and \textbf{+2.3 MAP@5}, with gains concentrated on layout-intensive pages. 
Our model also performs on par with or surpasses the Cross-Context oracle without requiring descriptors at inference. 
This indicates that our method can capture global structural signals not merely as exploitable features, but as parameterized retrieval prior.

\paragraph{Contributions.}
Our key contributions are
\textbf{(1)} We identify a limitation of late-interaction retrieval for visual documents, where the MaxSim lacks a mechanism to model page-level structure;
\textbf{(2)} We show that late-interaction retrieval fails to represent cross-region structure; so we introduce a learnable global bottleneck to encode it,
\textbf{(3)} We show that this training-only structural signal improves retrieval without requiring descriptors at inference, yielding consistent gains over comparable \colpali/\texttt{ColQwen} baselines, especially on layout-rich pages.

\begin{figure*}[!t]
    \centering
    \includegraphics[width=1.0\linewidth, trim={1.6cm 3.2cm 1.6cm 3.2cm}, clip]{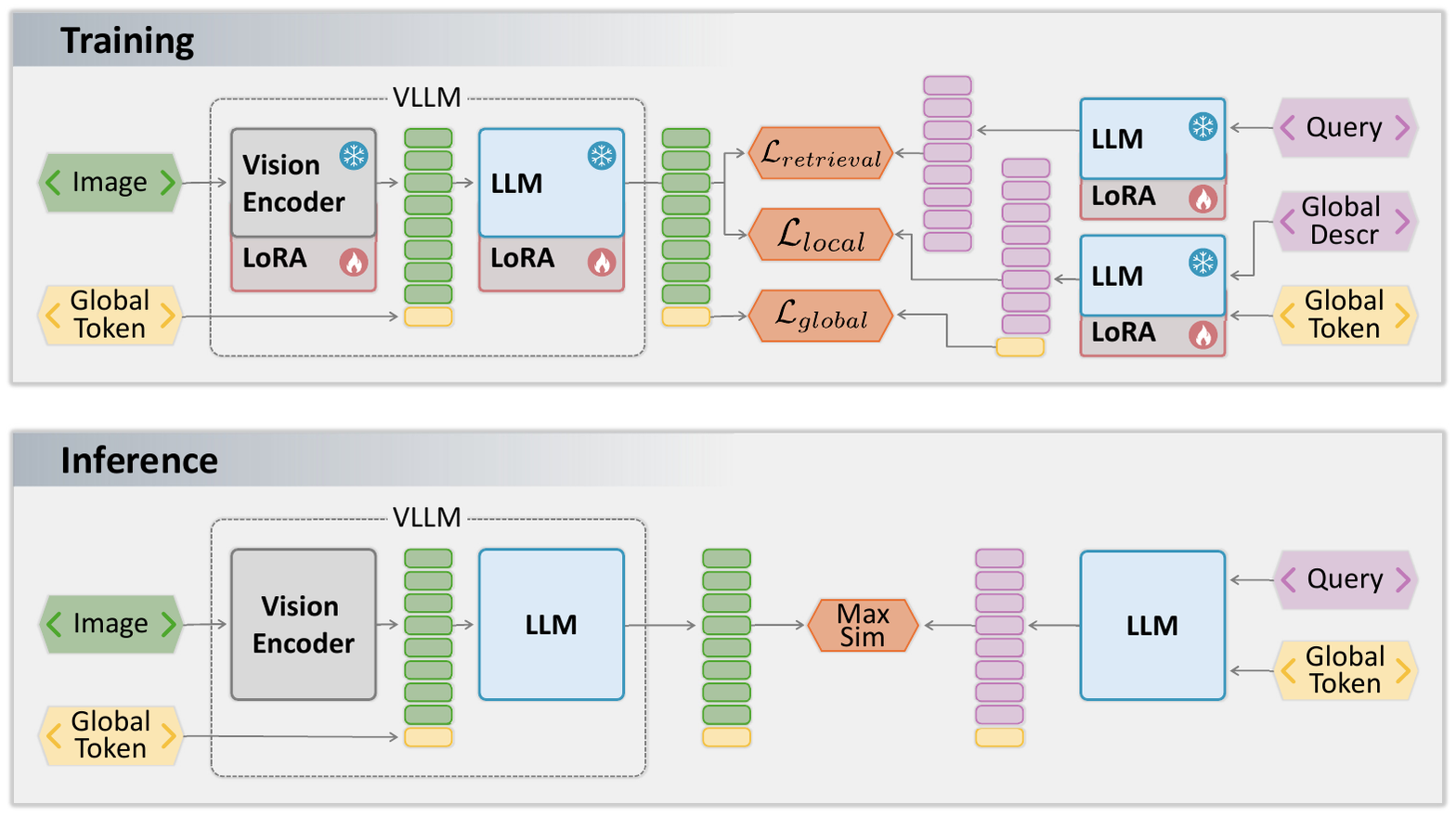}
    \caption{
    A high-level view of our descriptor-guided global modeling approach.
    \textbf{Top (Training):} We append a trainable global token to the vision-encoder patch embeddings and contextualize it jointly with local features in the \textsc{Vllm}. Training optimizes three objectives: (1) a late-interaction max-similarity loss between query and document patches, (2) a late-interaction max-similarity loss between textual global descriptors and document patches, and (3) a global-token InfoNCE alignment loss using cosine similarity logits aligning the image-derived global token with the descriptor-derived global representation.
    \textbf{Bottom (Inference):} Only image and query inputs are used. 
    The global token is appended to both sequences and contextualized without requiring textual descriptors. Document relevance is scored using late-interaction similarity.
    }
    \label{fig:method-global-description}
\end{figure*}

\section{Method}
\label{sec:method}

The late-interaction retrieval lacks a mechanism to encode global layout, which is a key signal for robust retrieval from visually-rich documents. 
This structural limitation is because patch embeddings are optimized independently and matched locally. 
We address this by introducing a global structural bottleneck.
This bottleneck enables page-level context to influence retrieval.
This context includes layout, spatial organization, and distributed visual patterns that cannot be reliably captured by independent patch embeddings nor by their simple mathematical aggregations (e.g., the mean pooling) as we show in our experiments. 

\Cref{fig:method-global-description} shows an overview of our method.
We introduce a multi-vector visual encoder augmented with a trainable global token and descriptor-guided structural supervision. 
The global token acts as a dedicated aggregation variable using self-attention. 
By attending to all patches simultaneously, the global token encodes the relative spatial relationships, such as the proximity of a caption to a chart. 
These relationships are otherwise lost in independent local patch matching. 
This consolidates page-level layout information into a singular representation rather than leaving it implicitly distributed across patches. 


\subsection{Training}
\paragraph{Multi‑Vector Encoder with Trainable Global Token.}
We build on a \gls*{vllm} that projects a document page into a sequence of patch embeddings
$ \mathbf{I} \in \mathbb{R}^{L_p \times d} $. 
While the transformer backbone allows for inter-patch attention during encoding, the late-interaction similarity function matched at retrieval time operates on patches independently, effectively treating the page as a ``bag-of-patches.'' 
To preserve and aggregate global page‑level semantics, we append a trainable $\mathbf{g_v} \in \mathbb{R}^d$ token to the visual sequence  
$ \mathbf{I}' = [\, \mathbf{I}; \mathbf{g_v} ]\ \in \mathbb{R}^{(L_p + 1) \times d}$, which is randomly initialized.
By passing $\mathbf{I}'$ through the self-attention layers of the language model, $\mathbf{g_v}$ acts as a structural bottleneck. 
It is trained to aggregate cross-region relationships that would otherwise be lost during the independent MaxSim matching phase. 
As a result, global layout information is explicitly represented in the multi-vector index and can influence retrieval alongside local patch embeddings. 

\paragraph{Textual Structural Descriptors as Layout Supervision.}
To train the global token $\mathbf{g}$ to represent layout priors, we supervise it using textual structural descriptors.
These descriptors (\Cref{fig:teaser}) summarize a page's spatial organization (e.g.,  the positioning of columns, tables, and headers) while omitting fine-grained textual content. 
By abstracting away local details, we force the global token to encode global page layout rather than a redundant representation of local patches.
These descriptors are used only as an auxiliary signal during training. 
At inference, the model operates as a purely visual retriever.
Further details about generation of these descriptors are in Appendix \ref{app:gbl-text}. 

\paragraph{Aligning Global Token to Textual Descriptions.}
We enforce structural consistency at the page level by aligning the visual global token with the descriptor's textual representation. 
The textual descriptor is passed through the same \gls*{llm} backbone of the \gls*{vllm}.
It ensures that the text encoder and vision-language encoder share the same semantic vocabulary and places both visual and textual tokens in a shared embedding space. 
The embedding layer of the LLM encodes the textual descriptor into a sequence $\mathbf{E}_{\mathrm{desc}} \in \mathbb{R}^{L_T \times d}$, where $L_T$ is the number of textual tokens in $\mathrm{desc}$. 
We randomly initialized a global token embeddings $\mathbf{g}\in \mathbb{R}^{d}$ and append it to $\mathbf{E}_{\mathrm{desc}}$. 
After processing through the self-attention layers, this token's output hidden state serves as the textual structural token $\mathbf{g}_{\mathrm{desc}} \in \mathbb{R}^{d}$.

We apply the InfoNCE objective, which is a contrastive loss, averaged over the batch, to align $\mathbf{g}$ with $\mathbf{g}_{\mathrm{desc}}$: 
\begin{equation}
\mathcal{L}_{\mathrm{global}}
= -\frac{1}{B}\sum_{i=1}^{B}
\left(
s_{ii} - \log \sum_{j=1}^{B} \exp(s_{ij})
\label{eq:global-loss}
\right),
\end{equation}

where $s_{ij} = \frac{\cos(\mathbf{g}_v^i, \mathbf{g}_{\mathrm{desc}}^j)}{\tau}$, $\mathbf{g}_v^i$ and $\mathbf{g}_{desc}^i$ denote the visual and textual global token of the $i$-th sample in the batch, respectively.
$\tau$ is a temperature hyper-parameter and $B$ is the batch size. 
By using in-batch negatives, the global token learns the unique layout semantics of a page relative to other pages.
Unlike standard image-caption matching, this loss trains the visual global token to serve as a latent proxy for the high-level layout priors, effectively distilling the descriptor's structural semantics into the visual embedding space. 
%

\paragraph{Aligning Textual Descriptors with Local Patch Embeddings.}
To complement the global token’s macro-level awareness, we enforce fine-grained structural alignment at the patch level. 
We optimize local patch embeddings $\mathbf{I} \in \mathbb{R}^{L_p \times d}$ to align with the textual descriptor embeddings $\mathbf{E}_{\mathrm{desc}} \in \mathbb{R}^{L_g \times d}$:   
\begin{equation}
\mathcal{L}_{\mathrm{local}} = - \sum_{k=1}^{L_p} \max_{1 \le j \le L_g} \cos(\mathbf{i}_k, \mathbf{e}_j),
\end{equation}
%
which encourages each patch to match the most relevant structural descriptor token. 
By grounding local patches in the structural context provided by the descriptor, we ensure the model maintains structural coherence even when queried at a granular level. 

\paragraph{Joint Objective for Global and Local Structured Retrieval.}
We combine our structural alignment objectives with the standard multi-vector retrieval loss $\mathcal{L}_{\mathrm{retrieval}}$, which aligns patch embeddings $\mathbf{I}$ directly with query embeddings $\mathbf{Q}$ via late interaction.  
Thus, the final joint objective is:
\begin{equation}
\mathcal{L} = \mathcal{L}_{\mathrm{global}} + \mathcal{L}_{\mathrm{local}} +  \mathcal{L}_{\mathrm{retrieval}}
\end{equation}
This objective couples global structure with local evidence, enabling both to influence the learned representation space. 
To maintain computational efficiency, we restrict updates to LoRA adapters within the VLLM transformer layers, keeping the base vision encoder and LLM weights frozen. 
By coupling local evidence with global structure, the model learns a multi-granularity representation space where both levels contribute to the final retrieval score. 
The auxiliary objectives introduce negligible overhead, as they operate on existing hidden states via cosine similarity.

\subsection{Inference and Scoring}
At inference, our model operates as a purely visual retriever and does not require any textual descriptors.  
This makes the method  computationally viable for large-scale RAG settings where real-time descriptor generation would be prohibitively expensive.
Both queries and documents are encoded as multi-vector representations augmented with global tokens:
$\mathbf{Q} \in \mathbb{R}^{(L_q+1) \times d}$ and $\mathbf{D} \in \mathbb{R}^{(L_d+1) \times d}$. 
We compute the retrieval score using the late-interaction function:
\begin{equation}
s(\mathbf{Q}, \mathbf{D}) = \sum_{i=1}^{L_q+1} \max_{1 \le j \le L_d+1} \mathbf{Q}_i \mathbf{D}_j^\top,
\end{equation}
where $\mathbf{Q}$ and $\mathbf{D}$ are the query and document embedding sequences (including global tokens).

Including the global tokens in the MaxSim operation allows the model to consider both local matches and global structure.
Local patch similarities capture fine-grained evidence.
The global token captures coarse layout structure.
Together, they influence the final similarity score.
This mechanism allows structural priors learned from textual supervision to directly influence the similarity score.
It does so without increasing computational complexity beyond the addition of a single vector per document.
\section{Experiments}
\label{sec:experiments}

\begin{table}[!b]
    \small
    \centering
    \begin{tabular}{@{}lcccc@{}}
    \toprule  
    \textbf{Dataset} & \textbf{Docs} & \textbf{Queries} & \textbf{Pages} & \textbf{Pages / Query} \\  
    \midrule
    Economics & 5 & 232 & 452 & 15.6 \\  
    Biomedical & 27 & 640 & 1,016 & 3.2 \\  
    ESGH & 27 & 52 & 1,538 & 2.5 \\ 
    ESGR & 30 & 228 & 1,538 & 3.9 \\  
    \bottomrule
    \end{tabular}  
    \caption{The dataset statistics of \gls*{vidorev2}.}  
    \label{tab:vidore_datasets}  
\end{table}

\subsection{Datasets}
We evaluate \gbldescft{} (Section~\ref{sec:method}) on \gls*{vidorev2} \cite{vidorev2}, the current standard for assessing visual retrieval models under realistic document layouts. 
\Cref{tab:vidore_datasets} summarizes the dataset statistics.
We evaluate on four diverse domains: economic reports, biomedical materials, and ESG industry disclosures (ESGH and ESGR). 
These datasets exhibit heterogeneous structures, including multi-column text, embedded charts, and complex tabular data.

For fine-tuning, we use the training set of \colpali~\cite{colpali}, consisting of 118k training and 500 development samples. 
Each instance pairs an English text query with a relevant document page.
We augment each instance with a textual structural descriptor, as a one-time offline preprocessing step, (Appendix~\ref{app:gbl-text}) to provide supervision for global layout.
This setup enables controlled evaluation of retrieval under diverse layout conditions.

\subsection{Evaluation Metrics}
We report \gls*{ndcg} and \gls*{map}. 
\gls*{ndcg}\texttt{@k} measures ranking quality by penalizing relevant documents that appear lower in the results list. 
This is critical for RAG, where top-ranked pages dominate generation quality. 
\gls*{map}\texttt{@k} measures retrieval consistency by averaging precision across all relevant pages for a given query.
It captures a model’s ability to retrieve all relevant pages even in use cases with long-tail queries.  
For all experiments, we report metrics at $k=5$, following the default setting established by the ViDoRe benchmark.

\subsection{Methods Compared}
To study the impact of structural supervision, we compare \gbldescft{} with two architectural variants and state-of-the-art visual retrievers.

\begin{figure}[!t]
    \small
    \centering
    \includegraphics[width=1.0\linewidth, trim={0.2cm 5.4cm 0.2cm 5.4cm}, clip]{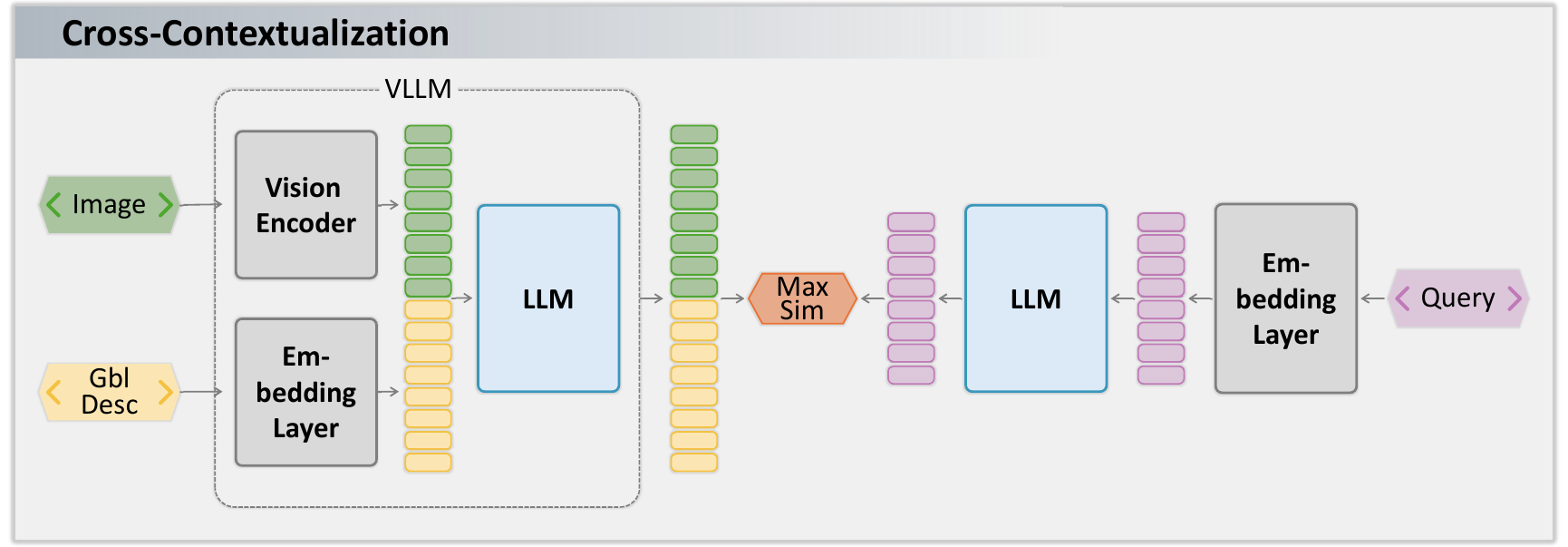}
    \caption{
    The cross-context baseline for frozen and fine-tuned variants.
    Patch-level image embeddings are concatenated with embeddings of tokens of the textual global descriptor.
    }
    \label{fig:cross-context}
\end{figure}

\paragraph{Cross-Context-Frozen  (\cczeroshot{}).}
We construct a training-free variant that concatenates local visual patches with pre-generated textual descriptors at inference time. 
Visual and textual embeddings are concatenated and passed through the~\gls*{vllm}, allowing joint attention over local details and page layout semantics (\Cref{fig:cross-context}).
This training-free variant serves as an oracle baseline as it uses the generated textual descriptors during the retrieval phase. 
While computationally expensive, it provides an insight on to what extend textual descriptor could be informative for retrieval. 

\paragraph{Cross-Context-FineTuned (\ccft{}).}
To examine whether global structure can be encoded by the model parameters when textual structural descriptors are available, we fine-tune the cross-context variant on the same \texttt{vidore/colpali\_train\_set} used for \gbldescft{} training. 
Each training instance consists of a query, its relevant document page, and the offline-generated textual structural descriptor for that page. 
The descriptor tokens are concatenated with the visual patch tokens and passed through the \gls*{vllm}, enabling joint contextualization of local visual evidence and page-level descriptor information. 
We update only \gls*{lora} parameters, keeping all base model weights frozen.
The model is optimized with the standard late-interaction retrieval objective over query--document pairs using in-batch negatives, and early stopping is performed on the development split. 

\begin{table*}[ht]
    \small
    \centering
    \begin{tabular}{@{}lr cccccccc |cc@{}} 
        \toprule
         & 
         &
        \multicolumn{2}{c}{\textbf{Economics}} & 
        \multicolumn{2}{c}{\textbf{Biomedical}} & 
        \multicolumn{2}{c}{\textbf{ESGH}} & 
        \multicolumn{2}{c}{\textbf{ESGR}} & 
        \multicolumn{2}{|c}{\textbf{Avg}}
        \\
        \textbf{Method} & 
        \textbf{Size} & 
        \textbf{\gls*{ndcg}} & \textbf{\gls*{map}} &
        \textbf{\gls*{ndcg}} & \textbf{\gls*{map}} &
        \textbf{\gls*{ndcg}} & \textbf{\gls*{map}} &
        \textbf{\gls*{ndcg}} & \textbf{\gls*{map}} &
        \textbf{\gls*{ndcg}} & \textbf{\gls*{map}} 
        \\
        \midrule
         \multicolumn{12}{@{}l}{\emph{Architecturally and size comparable}}
        \\
        \colpali~\cite{colpali} & 2B & 47.5 & 17.7 & 55.2 & 46.1 & 52.2 & 46.1 & 48.5 & 36.1 & 50.9 & 36.5 \\
        \colnomicthree~\cite{nomicembedmultimodal2025} & 3B & 54.3 & 22.5 & \textbf{62.6} & \textbf{53.2} & 64.3 & 57.9 & 56.4 & 43.5 & 59.4 & 44.3 \\
        \colqwen{}-\texttt{T} & 3B & 53.0 & 21.0 & 58.4 & 49.2 & 62.0 & 54.8 & 51.9 & 39.6 & 56.3 & 41.2\\
        \colqwen{}~\cite{colpali}            & 3B & 55.9 & 23.4 & 61.2 & 52.1 & 68.1 & 60.4 & 55.9 & 43.8 & 60.3 & 45.0 \\
        \gbldescft{} $_{\text{(ours)}}$    & 3B & \textbf{58.2}$^{\dagger}$ & \textbf{25.2}$^{\dagger}$ & 61.7$^{\dagger}$ & 52.4$^{\dagger}$ & \textbf{73.4}$^{\dagger}$ & \textbf{67.1}$^{\dagger}$ & \textbf{57.3}$^{\dagger}$ & \textbf{44.6}$^{\dagger}$ & \textbf{62.7} & \textbf{47.3}\\
        \midrule
          \multicolumn{12}{@{}l}{\emph{Architecturally different OR significantly larger}}\\
        \colnomicseven~\cite{nomicembedmultimodal2025} & 7B & 55.5 & 23.7 & 62.9 & 54.3 & 69.8 & 62.4 & 54.1 & 43.1 & 60.6 & 45.9 
        \\
        \nemothree~\cite{Nemoretriever}     & 4.4B & 57.8 & 24.4 & 62.7 & 53.9 & 75.4 & 68.3 & 57.6 & 45.5 & 63.5 & 48.0 \\
        \bottomrule
    \end{tabular}
    \caption{
    We report performance on \gls*{vidorev2} using nDCG@5, MAP@5, and their means.
    The first block contains architecturally comparable multi-vector retrievers with similar parameter scale (2–3B), representing our primary baselines.
    \gbldescft{} improves consistently over the strongest 3B \colpali{}-family baseline (\colqwen{}).
    The second block includes larger or architecturally different models (4–7B) trained with different objectives and is shown only for reference.
    ESGH and ESGR are the human-labeled and Restaurant version of ESG dataset.
    \colqwen{}-\texttt{T} is \colqwen{} where supplied with OCR-based textual representations of documents. 
    Results marked with ${\dagger}$ are statistically significant (Wilcoxon Signed-Rank test, $p < 0.05$) compared to the \colqwen{}. 
    }
    \label{tab:results-nDCG5-summary}  
\end{table*}

\paragraph{State-of-the-art (SOTA) Methods.}
We compare our method with leading open-source VDR systems on  \gls*{vidorev2}. 
Our primary baseline is \colqwen~\cite{colpali}, which shares the same architecture and pretraining as our model.
We additionally include \nemoretriever~\cite{Nemoretriever}, \colnomic{} \cite{nomicembedmultimodal2025},
and \colpali{} \cite{colpali}
to provide a comprehensive view of the current state of visual retrieval.
These models differ in pretraining objectives and data, and are evaluated without additional fine-tuning for comparability.

\subsection{Implementation Details}
The vision and text encoders are initialized from \colqwen{} pretrained \gls*{lora} adapters and weights. 
The global token embedding is initialized randomly and added to the model's vocabulary. 
Base encoder weights remain frozen; only \gls*{lora} parameters and the retrieval projection head are updated.
We fine-tune \gbldescft{} for 6~epochs with a batch size of 128 using AdamW with a learning rate of $5\times 10^{-5}$, weight decay $10^{-4}$, and a linear warm-up over the first 100 steps.
The full implementation details are in Appendix \ref{sec:implementation_details}. 
\section{Results}
\label{sec:results}
\paragraph{Main Findings.}
Overall, our evaluation reveals four primary trends that characterize the impact of global structural descriptors on retrieval performance: 
\textbf{(1)} \gbldescft{} consistently outperforms all architecturally comparable baselines. 
We observe a +2.4 point improvement in average nDCG@5 and a +2.3 point gain in average MAP@5 over our primary baseline, \colqwen{}.
\textbf{(2)} These gains concentrate on layout-rich datasets (e.g., ESGH), demonstrating the importance of page-level layout signals.
\textbf{(3)} Ablations reveal that global and local cues are individually insufficient and so their synergy drives the gains.
\textbf{(4)} Improvements scale with spatial heterogeneity and multi-region reasoning demands.
Together, these findings show that our descriptor-guided global alignment idea yields a structural inductive bias for visual document retrieval and cannot be recovered from patch-level similarity alone.

\begin{table*}[!th]  
    \small  
    \centering  
    \begin{tabular}{@{}l cccccccc |cc@{}}  
        \toprule  
        & \multicolumn{2}{c}{\textbf{Economics}} &  
        \multicolumn{2}{c}{\textbf{Biomedical}} &  
        \multicolumn{2}{c}{\textbf{ESGH}} &  
        \multicolumn{2}{c}{\textbf{ESGR}} &  
        \multicolumn{2}{|c}{\textbf{Avg}} \\  
        \textbf{Method} &  
        \textbf{\gls*{ndcg}} & \textbf{\gls*{map}} &  
        \textbf{\gls*{ndcg}} & \textbf{\gls*{map}} &  
        \textbf{\gls*{ndcg}} & \textbf{\gls*{map}} &  
        \textbf{\gls*{ndcg}} & \textbf{\gls*{map}} &  
        \textbf{\gls*{ndcg}} & \textbf{\gls*{map}} \\  
        \midrule  
        \gbldescft & 58.2 & 25.2 & 61.7 & 52.4 & 73.4 & 67.1 & 57.3 & 44.6 & \textbf{62.7} & \textbf{47.3} \\
        \hspace{3mm} w/o local patch embeddings  & 49.2 & 18.0 & 43.8 & 35.8 & 28.1 & 22.8 & 19.5 & 12.9 & 35.2 & 22.4 \\
        \hspace{3mm} w/o query global token & 58.1 & 24.9 & 62.2 & 52.9 & 71.6 & 65.0 & 57.7 & 45.3 & 62.4 & 46.9 \\
        \hspace{3mm} w/o image global token & 58.4 & 25.1 & 61.9 & 52.6 & 72.5 & 66.2 & 57.4 & 44.9 & 62.6 & 47.2 \\
        \midrule
        Fine-tuned w/o $\mathcal{L}_{\mathrm{global}}$ & 55.0 & 23.5 & 60.2 & 51.4 & 69.7 & 62.2 & 57.0 & 45.1 & 60.5 & 45.6 \\  
        Fine-tuned w/o $\mathcal{L}_{\mathrm{local}}$ & 57.8 & 24.9 & 60.9 & 52.0 & 71.0 & 64.1 & 57.5 & 45.1 & 61.8 & 46.5 \\  
        \midrule
        Mean-Pooling & 38.7 & 28.7 & 17.1 & 17.6 & 25.5 & 14.0 & 22.6 & 13.1 & 11.7 & 15.3 \\
        Max-Pooling & 16.3 & 13.6 & 6.6 & 10.2 & 11.7 & 4.0 & 9.9 & 4.6 & 7.7 & 6.5 \\
        Median-Pooling & 38.5 & 26.1 & 16.9 & 16.7 & 24.6 & 13.9 & 20.5 & 12.5 & 10.9 & 14.4 \\

        \bottomrule
    \end{tabular}  
    \caption{  
    Ablation study on ViDoRe-v2.  
    The Mean, Max, and Median poolings are deterministic ways to induce global representations from local patch embeddings. 
    }  
    \label{tab:ablation}  
\end{table*}

\subsection{Document Retrieval}
\Cref{tab:results-nDCG5-summary} reports retrieval effectiveness across the four \gls*{vidorev2} datasets. 

\gbldescft{}\ achieves the strongest average performance among all ColPali- and ColQwen-based systems, which are closest counterparts in terms of architecture and training setup. 
It gains +2.4 nDCG@5 and +2.3 MAP@5 points compared with \colqwen{}, showing that the proposed global layout modeling encodes a complementary structural prior not captured by patch-level similarity. 

When comparing with large multimodal retrievers such as \nemothree{} (4.4B) and ColNomic-7B; our 3B-parameter model delivers competitive average performance, and in some cases exceeds ColNomic-7B, in particular on layout-heavy datasets ESGH and ESGR. 
This suggests improved parameter efficiency as our 3B model outperforms a 7B model by leveraging structural inductive biases. 
%
Crucially, the underperformance of \text{\colqwen{}-T} suggests that raw text without spatial context provides limited benefit.
Together, these findings show that a compact model with explicit structural bias can match and even sometimes surpass the performance of substantially larger encoders on visually heterogeneous documents.

\subsection{Ablation Study}
\Cref{tab:ablation} contrasts the contribution of global and local patch embeddings as well as alternative pooling strategies for aggregating local patch embeddings to obtain global embeddings.  
while we report per-dataset scores for completeness, we focus our analysis on the mean (Avg) scores, which are more reliable than dataset-specific fluctuations.

Removing local patch embeddings leads to performance degradation on average (\gls*{ndcg}: $-27.5$, \gls*{map}: $-24.9$). 
In this configuration, the model relies solely on the global token embeddings, which encode page-level layout. 
Indeed, this model lacks the local content granularity that is essential to precisely match localized elements such as individual words, numbers, or icons. 
This confirms that global structure embeddings contributes to the retrieval performance and local patches are still the primary evidence for fine-grained matching.  
%
Ablating the global tokens leads to small but consistent declines (\gls*{ndcg}: $-0.3$, \gls*{map}: $-0.4$). 
The small magnitude of this drop suggests that the fine-tuned local patches have partially learned to capture the structural signals of document. 
The explicit token still provides the optimal anchor. 
It adds information that improves performance on layout-heavy pages.
%
Ablating either the query- or image-side global token yields similar drops (\gls*{ndcg}$\leq-0.4$).
This symmetry suggests that the benefit of global structure comes from aligning query intent with document layout, not from either side alone.
The largest gains are when both are present, enabling joint reasoning over query intent and document structure.

\paragraph{Effects of the Loss Functions}
Removing the global-token alignment loss (w/o $L_{\text{global}}$) drops the performance  (\gls*{ndcg}: $-2.2$, \gls*{map}: $-1.7$). 
This shows that explicit supervision for training the global representation is the larger contributor in this ablation.
Dropping the local-descriptor alignment loss (w/o $L_{\text{local}}$) also lowers performance (\gls*{ndcg}: $-0.9$, \gls*{map}: $-0.8$), indicating that patch-level structural grounding is essential for retrieval as well. 

\paragraph{Pooling is an Inadequate Substitute.} 
In the Mean, Max, and Median pooling variants of \gbldescft, we use these aggregation functions to obtain global token representations from local patch embeddings. 
We observe a substantial performance drop, confirming that naively collapsing local patch information into a single global vector removes crucial structure signals and cannot replace trained global representations by \gbldescft.

\subsection{Impact of Global Descriptors} 
\begin{table}[!t]
    \small
    \centering
    \begin{tabular}{lcc}
    \toprule
    \textbf{Method} & \textbf{\gls*{ndcg}} & \textbf{\gls*{map}}\\
    \midrule
         \cczeroshot & 61.7& 46.4 \\
         \ccft       & 62.5 &	47.1 \\
         \gbldescft  &  \textbf{62.7} & \textbf{47.3} \\
         \bottomrule
    \end{tabular}
    \caption{Comparing \gbldescft{} with \cczeroshot{} and \ccft{} using mean nDCG and MAP scores on ViDoRe-v2.  }
    \label{tab:global-descriptions}
\end{table}

\Cref{tab:global-descriptions} compares \gbldescft{} to its two \text{Cross-Context} counterparts (\cczeroshot{} and \ccft{}).
These models require textual global descriptors at inference.
Despite not using descriptors at inference, our \gbldescft{} outperforms \cczeroshot{} and performs on par with \ccft{}, indicating that page layout information extraction is learned via encoder parameters. 

\subsection{Case Study: Layout Complexity in ESGH}
\label{sec:app-case-study}
\begin{figure}
    \centering
    \includegraphics[width=1.0\linewidth,trim={1.7cm 3.5cm 1.7cm 3.5cm},clip]{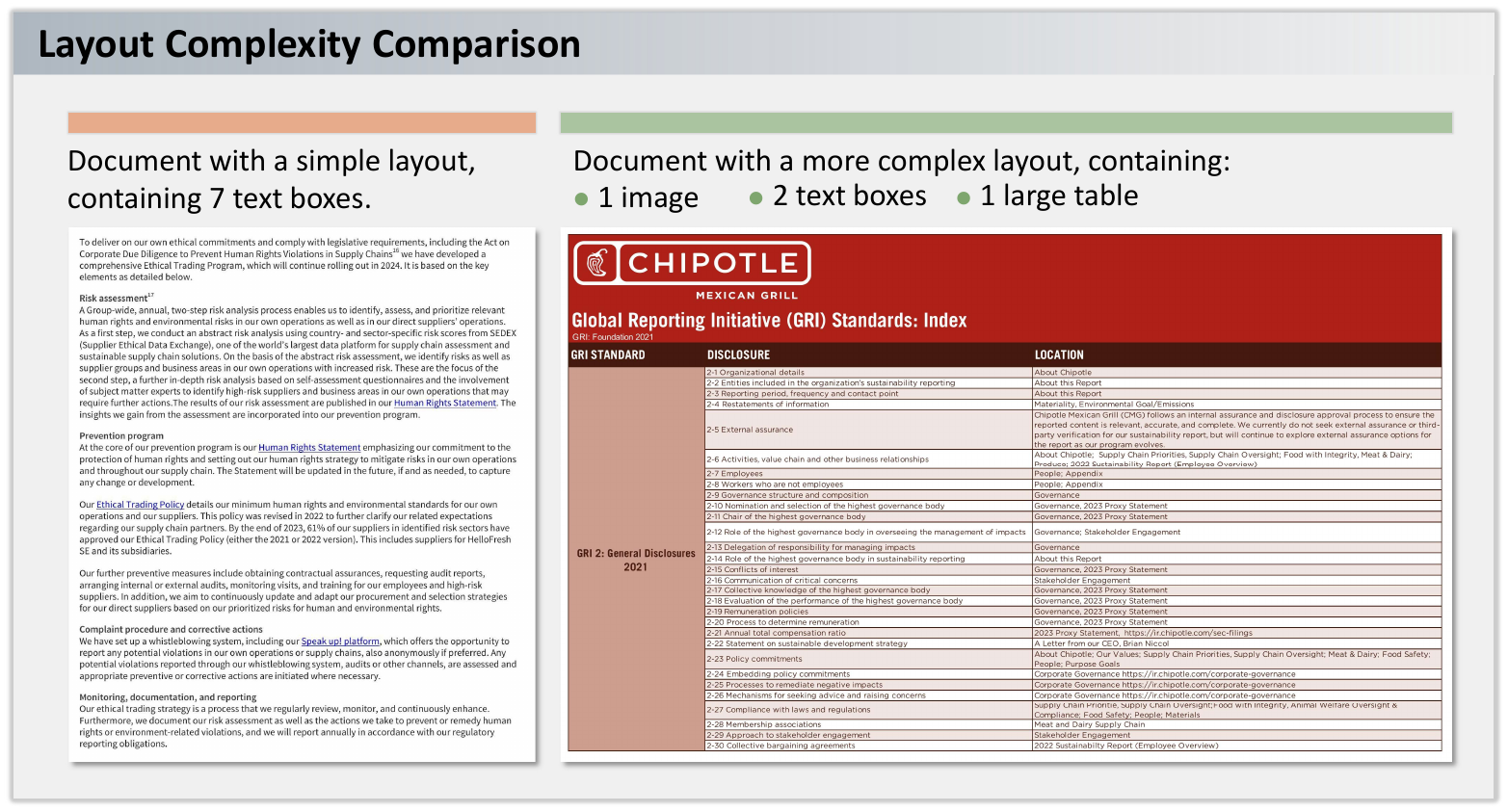}
    \caption{Examples of ESGH page layouts: a simple, text‑only page (top) and a complex, multi‑region layout with mixed modalities (bottom).}
    \label{fig:case-study}
\end{figure}

We use ESGH as a focused case study because it contains the most structurally complex layouts in \gls*{vidorev2}, including dense text regions, visual elements, lists, and spatially dispersed content, as illustrated in \Cref{fig:case-study} and summarized in \Cref{tab:failure_case_esgh}.   
\begin{table}[!t]
    \centering  
    \small
    \label{tab:failure_case_esgh}  
    \begin{tabular}{lrr}  
    \toprule  
    \textbf{Feature} & \textbf{Failures} & \textbf{Improvements} \\  
    \midrule  
    Text  & 11.56 & 14.14 \\  
    Image & 2.91 & 3.31 \\  
    List  & 1.47 & 2.80 \\ 
    \hline
    $N$ Words & 289.27 & 380.53 \\  
    $N$ VE & 2.24 & 3.24 \\  
    \bottomrule
    \end{tabular}  
    \caption{Summary of the mean of certain document page features of failure cases and improvements of our method compared to its baseline \colqwen{} on the ESGH dataset
    }
\end{table}

As shown in \Cref{fig:case-study}, simple pages with contiguous text boxes offer limited visual or spatial variation, whereas complex layouts combine images, multiple text regions, and large tables, requiring integration of visual semantics, textual cues, and layout relationships.
\Cref{tab:failure_case_esgh} shows that improvement cases consistently have more text regions, visual elements, lists, and higher content density than failure cases, reflecting the same complexity signals—high spatial entropy and element diversity—linked to larger gains in earlier analysis.

These patterns align with our quantitative findings: \gbldescft{} excels when spatial-semantic reasoning across heterogeneous, multi‑region layouts is needed, but offers smaller gains on simpler, text‑only pages where local matching suffices.
This case study thus confirms that descriptor-guided global modeling delivers its strongest benefits in visually rich, structurally complex documents.

\section{Related Work}
\label{sec:rel}

Recent visual document retrieval methods can be grouped into two broad categories, OCR-based retrieval and visual document retrieval.

\paragraph{OCR-based Retrieval}
Recent OCR-based methods \cite{wang-etal-2024-improving-text, deepseek-ocr, wei2024generalocrtheoryocr20, mineru25} typically follow two stages.
First they convert pages into textual and structural tokens, such as words, lines, tables, and detected layout regions, using an OCR engine. 
Retrieval is then performed in the text space with methods such as dense text encoders ~\cite{reimers-gurevych-2019-sentence}, BM25~\cite{bm25}, and layout-aware models~\cite{layoutlm, xu-etal-2021-layoutlmv2, layoutlmv3}. 
While effective, these pipelines inherit OCR errors such as incorrect reading order, misaligned bounding boxes, missing captions, or failure to parse figures and charts.
As a result, their performance degrades on visually rich pages, where crucial semantic cues are encoded in the raw layout, not in transcribed text.
In contrast, our approach operates directly on the visual document image and learns a global structural representation that does not depend on OCR or textual content. 

\paragraph{Visual Document Retrieval}

Visual document retrieval bypasses OCR by embedding an entire document page as an image and learning similarity directly in a multimodal space.
Early vision-language document models such as LayoutLM \cite{layoutlm, layoutlmv3} combine textual, layout, and visual cues for document understanding, but they are primarily designed for single-vector prediction settings rather than fine-grained retrieval.

Recent systems adopt a multi-vector late-interaction paradigm \cite{colpali, Nemoretriever, nomicembedmultimodal2025, gunther2025jina}, in which a page is represented as a \emph{set} of patch embeddings and relevance is computed by matching query tokens against the strongest-scoring page patches.
This formulation is effective for local cues such as words, table cells, or icons, but it biases scoring toward isolated matches.
In multimodal pages, however, relevance often depends on page-level structure, namely how textual and visual elements are arranged and related across the page.
Although some of this information may be present implicitly in visual encoders \cite{globalcitation}, it is not explicitly supervised and may be under-emphasized by late-interaction scoring.
Our work addresses this limitation by introducing descriptor-guided structural supervision for late-interaction retrieval, implemented through an explicit global token trained jointly with local patch embeddings.

Unlike global-pooling or CLS-token variants \cite{clip, layoutlmv3, dosovitskiy2021imageworth16x16words, zhai2022litzeroshottransferlockedimage}, our contribution is not the use of a single additional vector by itself.
Instead, we train a retrieval-aware global representation with textual holistic page descriptors that summarize layout and organization during training only.
This supervision encourages the encoder to represent cross-region structure while preserving the fine-grained local matching behavior of multi-vector retrieval, and it removes the need for textual descriptors at inference.

\section{Conclusions}
\label{sec:conclusions}
We proposed \gbldescft{}, a visual retrieval model that augments patch-based encoders with a learned global token trained via textual descriptor-guided supervision. 
Our model addresses a core limitation of late-interaction visual retrievers, which fail to capture structural relevance beyond local matches. 
On \gls*{vidorev2}, our method achieves significant gains over strong baselines with fewer parameters.
We observe the largest improvements on visually rich and multi-region documents. 
Ablations confirm that global context complements local features. 
Overall, our findings show that structural signals can be learned as part of the retrieval representation itself, enabling more robust and scalable document retrieval across diverse layouts.
Future work may extend our method beyond single-page retrieval to multi-page reasoning.

\bibliography{latex/custom}

@misc{zhai2022litzeroshottransferlockedimage,
      title={LiT: Zero-Shot Transfer with Locked-image text Tuning}, 
      author={Xiaohua Zhai and Xiao Wang and Basil Mustafa and Andreas Steiner and Daniel Keysers and Alexander Kolesnikov and Lucas Beyer},
      year={2022},
      eprint={2111.07991},
      archivePrefix={arXiv},
      primaryClass={cs.CV},
      url={https://arxiv.org/abs/2111.07991}, 
}

@misc{dosovitskiy2021imageworth16x16words,
      title={An Image is Worth 16x16 Words: Transformers for Image Recognition at Scale}, 
      author={Alexey Dosovitskiy and Lucas Beyer and Alexander Kolesnikov and Dirk Weissenborn and Xiaohua Zhai and Thomas Unterthiner and Mostafa Dehghani and Matthias Minderer and Georg Heigold and Sylvain Gelly and Jakob Uszkoreit and Neil Houlsby},
      year={2021},
      eprint={2010.11929},
      archivePrefix={arXiv},
      primaryClass={cs.CV},
      url={https://arxiv.org/abs/2010.11929}, 
}

@article{vidorev2,
  title={ViDoRe Benchmark V2: Raising the Bar for Visual Retrieval},
  author={Mac{\'e}, Quentin and Loison, Ant{\'o}nio and Faysse, Manuel},
  journal={arXiv preprint arXiv:2505.17166},
  year={2025}
}

@inproceedings{
colpali,
title={ColPali: Efficient Document Retrieval with Vision Language Models},
author={Manuel Faysse and Hugues Sibille and Tony Wu and Bilel Omrani and Gautier Viaud and CELINE HUDELOT and Pierre Colombo},
booktitle={The Thirteenth International Conference on Learning Representations},
year={2025},
url={https://openreview.net/forum?id=ogjBpZ8uSi}
}

@inproceedings{colbert,
  title={Colbert: Efficient and effective passage search via contextualized late interaction over bert},
  author={Khattab, Omar and Zaharia, Matei},
  booktitle={Proceedings of the 43rd International ACM SIGIR conference on research and development in Information Retrieval},
  pages={39--48},
  year={2020}
}

@inproceedings{bm25,
author = {Robertson, Stephen and Walker, Steve and Jones, Susan and Hancock-Beaulieu, Micheline and Gatford, Mike},
year = {1994},
month = {01},
pages = {0-},
title = {Okapi at TREC-3.}
}

@inproceedings{clip,
  title={Learning transferable visual models from natural language supervision},
  author={Radford, Alec and Kim, Jong Wook and Hallacy, Chris and Ramesh, Aditya and Goh, Gabriel and Agarwal, Sandhini and Sastry, Girish and Askell, Amanda and Mishkin, Pamela and Clark, Jack and others},
  booktitle={International conference on machine learning},
  pages={8748--8763},
  year={2021},
  organization={PmLR}
}

@article{gunther2025jina,
  title={jina-embeddings-v4: Universal Embeddings for Multimodal Multilingual Retrieval},
  author={G{\"u}nther, Michael and Sturua, Saba and Akram, Mohammad Kalim and Mohr, Isabelle and Ungureanu, Andrei and Wang, Bo and Eslami, Sedigheh and Martens, Scott and Werk, Maximilian and Wang, Nan and others},
  journal={arXiv preprint arXiv:2506.18902},
  year={2025}
}

@article{Nemoretriever,
  title={Llama Nemoretriever Colembed: Top-Performing Text-Image Retrieval Model},
  author={Xu, Mengyao and Moreira, Gabriel and Ak, Ronay and Osmulski, Radek and Babakhin, Yauhen and Yu, Zhiding and Schifferer, Benedikt and Oldridge, Even},
  journal={arXiv preprint arXiv:2507.05513},
  year={2025}
}

@inproceedings{layoutlmv3,
  title={Layoutlmv3: Pre-training for document ai with unified text and image masking},
  author={Huang, Yupan and Lv, Tengchao and Cui, Lei and Lu, Yutong and Wei, Furu},
  booktitle={Proceedings of the 30th ACM international conference on multimedia},
  pages={4083--4091},
  year={2022}
}

@inproceedings{layoutlm,
  title={Layoutlm: Pre-training of text and layout for document image understanding},
  author={Xu, Yiheng and Li, Minghao and Cui, Lei and Huang, Shaohan and Wei, Furu and Zhou, Ming},
  booktitle={Proceedings of the 26th ACM SIGKDD international conference on knowledge discovery \& data mining},
  pages={1192--1200},
  year={2020}
}

@article{rag_survey,
  title={A survey of multimodal retrieval-augmented generation},
  author={Mei, Lang and Mo, Siyu and Yang, Zhihan and Chen, Chong},
  journal={arXiv preprint arXiv:2504.08748},
  year={2025}
}

@inproceedings{lora,
title={Lo{RA}: Low-Rank Adaptation of Large Language Models},
author={Edward J Hu and yelong shen and Phillip Wallis and Zeyuan Allen-Zhu and Yuanzhi Li and Shean Wang and Lu Wang and Weizhu Chen},
booktitle={International Conference on Learning Representations},
year={2022},
url={https://openreview.net/forum?id=nZeVKeeFYf9}
}

@misc{nomicembedmultimodal2025,
  title={Nomic Embed Multimodal: Interleaved Text, Image, and Screenshots for Visual Document Retrieval},
  author={Nomic Team},
  year={2025},
  publisher={Nomic AI},
  url={https://nomic.ai/blog/posts/nomic-embed-multimodal},
}

@misc{jina-colembeds,
      title={jina-embeddings-v4: Universal Embeddings for Multimodal Multilingual Retrieval}, 
      author={Michael Günther and Saba Sturua and Mohammad Kalim Akram and Isabelle Mohr and Andrei Ungureanu and Sedigheh Eslami and Scott Martens and Bo Wang and Nan Wang and Han Xiao},
      year={2025},
      eprint={2506.18902},
      archivePrefix={arXiv},
      primaryClass={cs.AI},
      url={https://arxiv.org/abs/2506.18902}, 
}

@inproceedings{rag,
 author = {Lewis, Patrick and Perez, Ethan and Piktus, Aleksandra and Petroni, Fabio and Karpukhin, Vladimir and Goyal, Naman and K\"{u}ttler, Heinrich and Lewis, Mike and Yih, Wen-tau and Rockt\"{a}schel, Tim and Riedel, Sebastian and Kiela, Douwe},
 booktitle = {Advances in Neural Information Processing Systems},
 editor = {H. Larochelle and M. Ranzato and R. Hadsell and M.F. Balcan and H. Lin},
 pages = {9459--9474},
 publisher = {Curran Associates, Inc.},
 title = {Retrieval-Augmented Generation for Knowledge-Intensive NLP Tasks},
 url = {https://proceedings.neurips.cc/paper_files/paper/2020/file/6b493230205f780e1bc26945df7481e5-Paper.pdf},
 volume = {33},
 year = {2020}
}

@article{mmdocir,
  title={Mmdocir: Benchmarking multi-modal retrieval for long documents},
  author={Dong, Kuicai and Chang, Yujing and Goh, Xin Deik and Li, Dexun and Tang, Ruiming and Liu, Yong},
  journal={arXiv preprint arXiv:2501.08828},
  year={2025}
}

@article{unidoc,
  title={UNIDOC-BENCH: A Unified Benchmark for Document-Centric Multimodal RAG},
  author={Peng, Xiangyu and Qin, Cab and Chen, Zeyuan and Xu, Ran and Xiong, Caiming and Wu, Chien-Sheng},
  journal={arXiv preprint arXiv:2510.03663},
  year={2025}
}

@article{deepseek-ocr,
  title={DeepSeek-OCR: Contexts Optical Compression},
  author={Wei, Haoran and Sun, Yaofeng and Li, Yukun},
  journal={arXiv preprint arXiv:2510.18234},
  year={2025}
}

@misc{wei2024generalocrtheoryocr20,
      title={General OCR Theory: Towards OCR-2.0 via a Unified End-to-end Model}, 
      author={Haoran Wei and Chenglong Liu and Jinyue Chen and Jia Wang and Lingyu Kong and Yanming Xu and Zheng Ge and Liang Zhao and Jianjian Sun and Yuang Peng and Chunrui Han and Xiangyu Zhang},
      year={2024},
      eprint={2409.01704},
      archivePrefix={arXiv},
      primaryClass={cs.CV},
      url={https://arxiv.org/abs/2409.01704}, 
}

@article{granite,
  title={Granite Vision: a lightweight, open-source multimodal model for enterprise Intelligence},
  author={Team, Granite Vision and Karlinsky, Leonid and Arbelle, Assaf and Daniels, Abraham and Nassar, Ahmed and Alfassi, Amit and Wu, Bo and Schwartz, Eli and Joshi, Dhiraj and Kondic, Jovana and others},
  journal={arXiv preprint arXiv:2502.09927},
  year={2025}
}

@inproceedings{pytorch,
author = {Ansel, Jason and Yang, Edward and He, Horace and Gimelshein, Natalia and Jain, Animesh and Voznesensky, Michael and Bao, Bin and Bell, Peter and Berard, David and Burovski, Evgeni and Chauhan, Geeta and Chourdia, Anjali and Constable, Will and Desmaison, Alban and DeVito, Zachary and Ellison, Elias and Feng, Will and Gong, Jiong and Gschwind, Michael and Hirsh, Brian and Huang, Sherlock and Kalambarkar, Kshiteej and Kirsch, Laurent and Lazos, Michael and Lezcano, Mario and Liang, Yanbo and Liang, Jason and Lu, Yinghai and Luk, C. K. and Maher, Bert and Pan, Yunjie and Puhrsch, Christian and Reso, Matthias and Saroufim, Mark and Siraichi, Marcos Yukio and Suk, Helen and Zhang, Shunting and Suo, Michael and Tillet, Phil and Zhao, Xu and Wang, Eikan and Zhou, Keren and Zou, Richard and Wang, Xiaodong and Mathews, Ajit and Wen, William and Chanan, Gregory and Wu, Peng and Chintala, Soumith},
title = {PyTorch 2: Faster Machine Learning Through Dynamic Python Bytecode Transformation and Graph Compilation},
year = {2024},
isbn = {9798400703850},
publisher = {Association for Computing Machinery},
address = {New York, NY, USA},
url = {https://doi.org/10.1145/3620665.3640366},
doi = {10.1145/3620665.3640366},
booktitle = {Proceedings of the 29th ACM International Conference on Architectural Support for Programming Languages and Operating Systems, Volume 2},
pages = {929–947},
numpages = {19},
location = {La Jolla, CA, USA},
series = {ASPLOS '24}
}

@inproceedings{hf-transformers,
    title = "Transformers: State-of-the-Art Natural Language Processing",
    author = "Thomas Wolf and Lysandre Debut and Victor Sanh and Julien Chaumond and Clement Delangue and Anthony Moi and Pierric Cistac and Tim Rault and Rémi Louf and Morgan Funtowicz and Joe Davison and Sam Shleifer and Patrick von Platen and Clara Ma and Yacine Jernite and Julien Plu and Canwen Xu and Teven Le Scao and Sylvain Gugger and Mariama Drame and Quentin Lhoest and Alexander M. Rush",
    booktitle = "Proceedings of the 2020 Conference on Empirical Methods in Natural Language Processing: System Demonstrations",
    month = oct,
    year = "2020",
    address = "Online",
    publisher = "Association for Computational Linguistics",
    url = "https://www.aclweb.org/anthology/2020.emnlp-demos.6",
    pages = "38--45"
}

@article{gradient-checkpointing,
  title={Training deep nets with sublinear memory cost},
  author={Chen, Tianqi and Xu, Bing and Zhang, Chiyuan and Guestrin, Carlos},
  journal={arXiv preprint arXiv:1604.06174},
  year={2016}
}

@INPROCEEDINGS{globalcitation,
  author={Maniparambil, Mayug and Akshulakov, Raiymbek and Dahou Djilali, Yasser Abdelaziz and Seddik, Mohamed EI Amine and Narayan, Sanath and Mangalam, Karttikeya and O'Connor, Noel E.},
  booktitle={2024 IEEE/CVF Conference on Computer Vision and Pattern Recognition (CVPR)}, 
  title={Do Vision and Language Encoders Represent the World Similarly?}, 
  year={2024},
  volume={},
  number={},
  pages={14334-14343},
  doi={10.1109/CVPR52733.2024.01359}}

@misc{mineru25,
      title={MinerU2.5: A Decoupled Vision-Language Model for Efficient High-Resolution Document Parsing}, 
      author={Junbo Niu and Zheng Liu and Zhuangcheng Gu and Bin Wang and Linke Ouyang and Zhiyuan Zhao and Tao Chu and Tianyao He and Fan Wu and Qintong Zhang and Zhenjiang Jin and others},
      year={2025},
      eprint={2509.22186},
      archivePrefix={arXiv},
      primaryClass={cs.CV},
      url={https://arxiv.org/abs/2509.22186}, 
}

@article{gpt5,
  title={Openai gpt-5 system card},
  author={Singh, Aaditya and Fry, Adam and Perelman, Adam and Tart, Adam and Ganesh, Adi and El-Kishky, Ahmed and McLaughlin, Aidan and Low, Aiden and Ostrow, AJ and Ananthram, Akhila and others},
  journal={arXiv preprint arXiv:2601.03267},
  year={2025}
}

@inproceedings{wu-etal-2025-doc,
    title = "Doc-React: Multi-page Heterogeneous Document Question-answering",
    author = "Wu, Junda  and
      Xia, Yu  and
      Yu, Tong  and
      Chen, Xiang  and
      Harsha, Sai Sree  and
      Maharaj, Akash V  and
      Zhang, Ruiyi  and
      Bursztyn, Victor  and
      Kim, Sungchul  and
      Rossi, Ryan A.  and
      McAuley, Julian  and
      Li, Yunyao  and
      Sinha, Ritwik",
    editor = "Che, Wanxiang  and
      Nabende, Joyce  and
      Shutova, Ekaterina  and
      Pilehvar, Mohammad Taher",
    booktitle = "Proceedings of the 63rd Annual Meeting of the Association for Computational Linguistics (Volume 2: Short Papers)",
    month = jul,
    year = "2025",
    address = "Vienna, Austria",
    publisher = "Association for Computational Linguistics",
    url = "https://aclanthology.org/2025.acl-short.6/",
    doi = "10.18653/v1/2025.acl-short.6",
    pages = "67--78",
    ISBN = "979-8-89176-252-7"
}

@inproceedings{abootorabi-etal-2025-ask,
    title = "Ask in Any Modality: A Comprehensive Survey on Multimodal Retrieval-Augmented Generation",
    author = "Abootorabi, Mohammad Mahdi  and
      Zobeiri, Amirhosein  and
      Dehghani, Mahdi  and
      Mohammadkhani, Mohammadali  and
      Mohammadi, Bardia  and
      Ghahroodi, Omid  and
      Baghshah, Mahdieh Soleymani  and
      Asgari, Ehsaneddin",
    editor = "Che, Wanxiang  and
      Nabende, Joyce  and
      Shutova, Ekaterina  and
      Pilehvar, Mohammad Taher",
    booktitle = "Findings of the Association for Computational Linguistics: ACL 2025",
    month = jul,
    year = "2025",
    address = "Vienna, Austria",
    publisher = "Association for Computational Linguistics",
    url = "https://aclanthology.org/2025.findings-acl.861/",
    doi = "10.18653/v1/2025.findings-acl.861",
    pages = "16776--16809",
    ISBN = "979-8-89176-256-5"
}

@inproceedings{wang-etal-2024-improving-text,
    title = "Improving Text Embeddings with Large Language Models",
    author = "Wang, Liang  and
      Yang, Nan  and
      Huang, Xiaolong  and
      Yang, Linjun  and
      Majumder, Rangan  and
      Wei, Furu",
    editor = "Ku, Lun-Wei  and
      Martins, Andre  and
      Srikumar, Vivek",
    booktitle = "Proceedings of the 62nd Annual Meeting of the Association for Computational Linguistics (Volume 1: Long Papers)",
    month = aug,
    year = "2024",
    address = "Bangkok, Thailand",
    publisher = "Association for Computational Linguistics",
    url = "https://aclanthology.org/2024.acl-long.642/",
    doi = "10.18653/v1/2024.acl-long.642",
    pages = "11897--11916"
}

@inproceedings{reimers-gurevych-2019-sentence,
    title = "Sentence-{BERT}: Sentence Embeddings using {S}iamese {BERT}-Networks",
    author = "Reimers, Nils  and
      Gurevych, Iryna",
    editor = "Inui, Kentaro  and
      Jiang, Jing  and
      Ng, Vincent  and
      Wan, Xiaojun",
    booktitle = "Proceedings of the 2019 Conference on Empirical Methods in Natural Language Processing and the 9th International Joint Conference on Natural Language Processing (EMNLP-IJCNLP)",
    month = nov,
    year = "2019",
    address = "Hong Kong, China",
    publisher = "Association for Computational Linguistics",
    url = "https://aclanthology.org/D19-1410/",
    doi = "10.18653/v1/D19-1410",
    pages = "3982--3992"
}

@inproceedings{xu-etal-2021-layoutlmv2,
    title = "{L}ayout{LM}v2: Multi-modal Pre-training for Visually-rich Document Understanding",
    author = "Xu, Yang  and
      Xu, Yiheng  and
      Lv, Tengchao  and
      Cui, Lei  and
      Wei, Furu  and
      Wang, Guoxin  and
      Lu, Yijuan  and
      Florencio, Dinei  and
      Zhang, Cha  and
      Che, Wanxiang  and
      Zhang, Min  and
      Zhou, Lidong",
    editor = "Zong, Chengqing  and
      Xia, Fei  and
      Li, Wenjie  and
      Navigli, Roberto",
    booktitle = "Proceedings of the 59th Annual Meeting of the Association for Computational Linguistics and the 11th International Joint Conference on Natural Language Processing (Volume 1: Long Papers)",
    month = aug,
    year = "2021",
    address = "Online",
    publisher = "Association for Computational Linguistics",
    url = "https://aclanthology.org/2021.acl-long.201/",
    doi = "10.18653/v1/2021.acl-long.201",
    pages = "2579--2591"
}

@String(CVPR= {IEEE Conf. Comput. Vis. Pattern Recog.})

@String(CVPR  = {CVPR})

\clearpage
\setcounter{page}{1}

\section{Appendix}
\label{sec:appendix}



\subsection{Prompt for Textual Baseline}
\label{app:text-image-rep}
The instructions provided to GPT-5 for generating textual summaries from images of document pages is displayed in the following. 
The prompt guides the model to produce plain‑text descriptions that capture the main ideas and relevant details, while avoiding verbatim transcription or formatting artifacts:

\texttt{
You are given an image representing a single page from a PDF document.
Your task is to produce a concise plain‑text summary of the content in the image. 
Do not include any formatting tokens such as bold text, bullet points, or other markup; the output should be plain text only. Read the visible text and summarize its main ideas, key points, and relevant details instead of transcribing it verbatim. 
If any text is unclear or partially visible, note that it is unreadable rather than guessing. For non‑text elements such as images, charts, or diagrams, briefly describe their content and placement.
}

\subsection{Global Textual Descriptors}
\label{app:gbl-text}
The instructions provided to GPT‑5 for generating high‑level textual descriptions of images are shown below. 
The prompt directs the model to focus on overall scene composition, spatial arrangement, and thematic content, avoiding fine‑grained details unless they dominate the image: 
 
\texttt{
You are given an image and must produce a concise yet informative textual description that captures its overall scene, structure, and high-level visual concepts. 
Focus on the broad composition, spatial arrangement, and thematic content that can be perceived without examining small details, describing the type of scene or setting, the general layout, dominant visual patterns, and the main purpose or activity depicted if relevant. 
Avoid mentioning small objects, text fragments, or fine-grained elements unless they dominate the entire image, and use clear, natural language that reflects the image’s global visual meaning.
}



\subsection{Implementation Details}
\label{sec:implementation_details}
%
The vision and text encoders are initialized from \colqwen{} pretrained \gls*{lora} adapters and weights. 
The new \texttt{[global]} token embedding is initialized randomly and added to the model's vocabulary. 
Base encoder weights remain frozen; only \gls*{lora} parameters and the retrieval projection head are updated.
We fine-tune \gbldescft{} for 6~epochs with a batch size of 128 using AdamW with a learning rate of $5\times 10^{-5}$, weight decay $10^{-4}$, and a linear warm-up over the first 100 steps.
\ccft{} is tuned with early stopping on the development set, converging after approximately 0.25 epochs, compared to 6 epochs for \gbldescft{}. 
This reflects task difficulty rather than overfitting: \ccft{} has direct access to text descriptors at inference time, while \gbldescft{} must internalize the same structural priors into the visual global token without textual input, a harder objective that requires more training to converge.
We train in \texttt{bfloat16} precision and enable gradient checkpointing \cite{gradient-checkpointing} to maximize memory efficiency on the GPU.
We conduct all training and evaluation on a single NVIDIA H200 GPU (141\,GB memory) using PyTorch \cite{pytorch} and the HuggingFace Transformers library \cite{hf-transformers}.
Our \gbldescft{} implementation extends the open-source \texttt{ColPali} framework~\cite{colpali}, adapting its multimodal retrieval pipeline to incorporate global textual descriptions alongside local visual tokens.
All baselines (\nemoretriever{}, \colnomic{}) are evaluated using their official HuggingFace implementations.
We set the temperature $\tau = 0.02$.
We generate the textual global descriptors using GPT-5~\cite{gpt5}.

\subsection{Performance by Query Type: Global layout vs. Local Detail Oriented}
\begin{figure}[!th]
    \centering
    \includegraphics[trim={0.5cm 0.2cm 1.5cm 1.85cm},clip,width=0.9\linewidth]{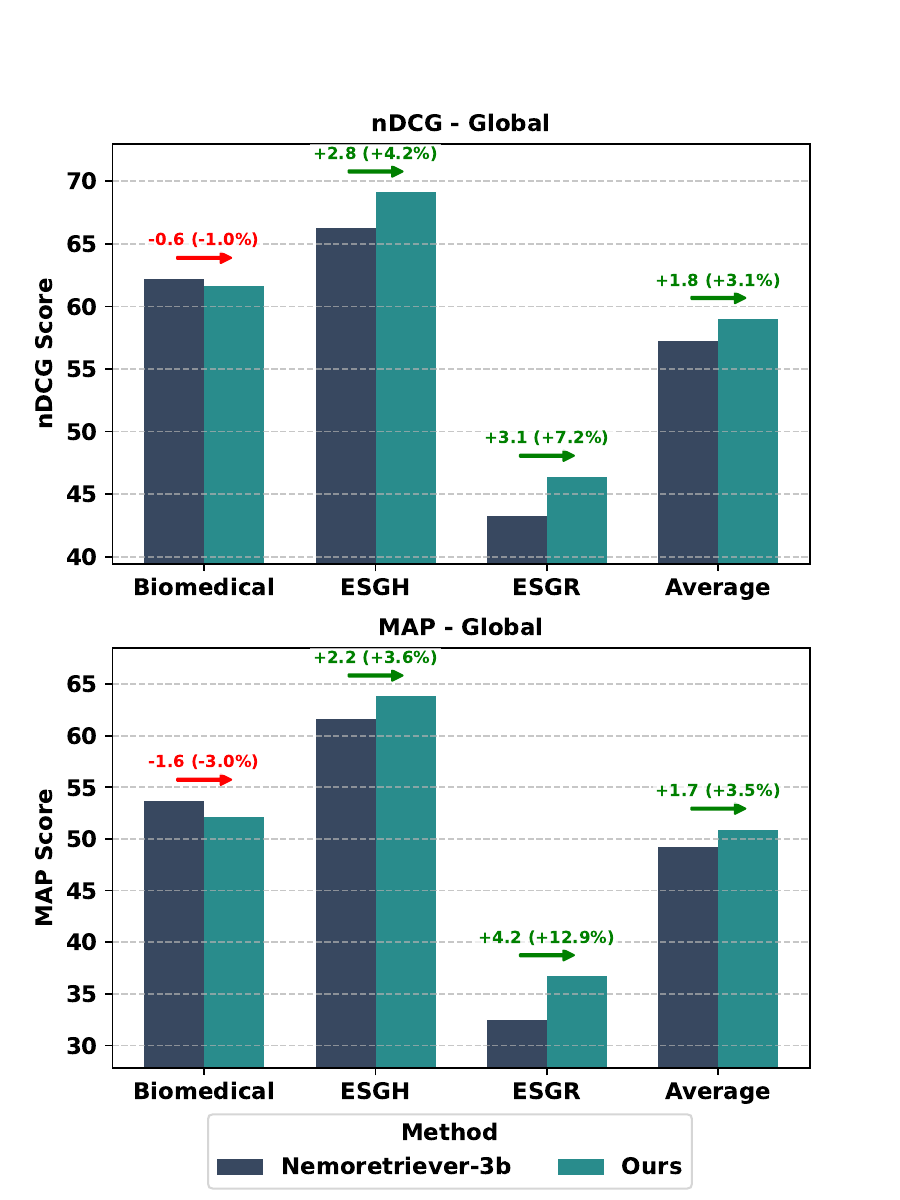}
    \caption{Performance on holistic‑context queries: Comparison of our \gbldescft{} model with \nemothree{} across Biomedical, ESGH, and ESGR datasets, and averaged results. Bars show \gls*{ndcg}\texttt{@5} (top) and \gls*{map}\texttt{@5} (bottom). Green arrows show absolute and relative increases of performance, red arrows show decreases.}
    \label{fig:global_performace}
\end{figure}
We label each query in \gls*{vidorev2} as either \emph{detail-oriented} or \emph{layout-oriented} using \gptfive{}. 

\Cref{fig:global_performace}  shows the performance of \gbldescft{} on layout oriented queries in comparison with \nemothree{}.  
The largest improvements belongs to the layout-rich ESGH and ESGR datasets, showing the soundness of the trained layout encoder.   
These two datasets contain multiple visual regions per page (tables, charts, and multi-column text), and relevance frequently depends on how these elements are spatially organized.  
In contrast, the Biomedical dataset is largely text dominant and contains many detail-oriented queries. 
Therefore, global layout information is less predictive for this dataset.
In such cases, \nemothree{}, which is optimized for localized visual cues, performs comparatively well.

\begin{table}[!t]  
    \small
    \centering  
    \begin{tabular}{lcc}  
    \toprule  
    \textbf{Dataset} & \textbf{Local} & \textbf{Global} \\  
    \midrule  
    Economics         & 2   & 230 \\  
    Biomedical        & 195 & 445 \\  
    ESG Human-Labeled & 41  & 11  \\  
    ESG Restaurant    & 128  & 100 \\ 
    \bottomrule  
    \end{tabular}
    \caption{The number of detail-oriented (\emph{local}) and layout-oriented (\emph{global}) queries per \gls*{vidorev2} dataset. 
    }  
    \label{tab:gbl_local_statistics}  
\end{table}

\subsection{Spatial Shannon Entropy}   
We quantify the spatial dispersion of layout elements by computing the Shannon entropy of their center-point distribution over a $G \times G$ grid.   
Let $\mathcal{B} = \{b_1, \dots, b_N\}$ be the set of detected blocks, and $\mathbf{c}_k = \left( \frac{x^{(k)}_{\min} + x^{(k)}_{\max}}{2}, \frac{y^{(k)}_{\min} + y^{(k)}_{\max}}{2} \right)$ their normalized center coordinates.   
Each $\mathbf{c}_k$ is assigned to a grid cell $(i,j)$, with counts $n_{ij}$ and cell probabilities $p_{ij} = n_{ij} / N$.   
The spatial Shannon entropy is then  
$  
H_{\text{spatial}} = - \sum_{i=1}^{G} \sum_{j=1}^{G} p_{ij} \log p_{ij},  
$  
where $p_{ij} = 0$ terms are omitted.   
Higher $H_{\text{spatial}}$ values indicate that elements are more uniformly distributed across the page, whereas lower values reflect spatial concentration.

\end{document}